\documentclass[10pt,twocolumn,letterpaper]{article}

\usepackage{cvpr}
\usepackage{times}
\usepackage{epsfig}
\usepackage{graphicx}
\usepackage{amsmath}
\usepackage{amssymb}
\usepackage{authblk}

\usepackage[breaklinks=true,bookmarks=false]{hyperref}

\cvprfinalcopy 

\def\cvprPaperID{3167} 

\begin{document}

\title{Loop Closure Detection with RGB-D Feature Pyramid Siamese Networks}

\author[2]{Zhang Qianhao} 
\author[3]{Alexander Mai}
\author[1]{Joseph Menke}
\author[1]{Allen Yang}
\affil[1]{University of California, Berkeley} 
\affil[2]{Beihang University} 
\affil[3]{University of California, San Diego}

\maketitle
\begin{abstract}
    In visual Simultaneous Localization And Mapping (SLAM), detecting loop closures has been an important but difficult task. Currently, most solutions are based on the bag-of-words approach. Yet the possibility of deep neural network application to this task has not been fully explored due to the lack of appropriate architecture design and of sufficient training data. In this paper we demonstrate the applicability of deep neural networks by addressing both issues. Specifically we show that a feature pyramid Siamese neural network can achieve state-of-the-art performance on pairwise loop closure detection. The network is trained and tested on large-scale RGB-D datasets with a novel automatic loop closure labeling algorithm. Each image pair is labelled by how much the images overlap, allowing loop closure to be computed directly rather than by labor intensive manual labeling. We present an algorithm to adopt any large-scale generic RGB-D dataset for use in training deep loop-closure networks. We show for the first time that deep neural networks are capable of detecting loop closures, and we provide a method for generating large-scale datasets for use in evaluating and training loop closure detectors.
\end{abstract}

\section{Introduction and Related Work}
Loop closure detection is the process of detecting whether an agent has returned to a previously visited location. This is critical for correcting accumulated errors over a large timescale in many real-world navigation applications. In a vision-based simultaneous localization and mapping (SLAM) system, loop closures are often detected via comparison of image pairs through the journey. 

Detecting loop closures has been a challenging task for its inherent susceptibility to variations within the scenes. A loop closure detection algorithm must be sensitive as to avoid classifying similar rooms as a positive detection while being resilient to small changes in object location, shifted viewpoints, different lighting conditions and shadows that could drastically alter the visual representation of the same scene.

\subsection{Bag-of-Words Approach}
The bag-of-words methodology was first proposed for text document analysis \cite{salton1983introduction} and was further adapted for computer vision applications \cite{bosch2007best}. For image analysis, a visual analogue of a word is used in the bag-of-words model, which is based on the vector quantization process by clustering low-level visual features of local regions or points, such as color, texture, and so forth \cite{tsai2012bag}.

Currently, bag-of-words approach is the state-of-the-art method for loop closure detection \cite{cummins2008fab,dbow2,mur2015orb,mur2017orb} in which each image is represented as a histogram of word-frequency of each word present in the dictionary generated offline from a large number of images. The computation of similarity is based on comparing the histograms \cite{zhang2010keyframe} between image pairs with certain heuristics such as spatial constraint or dynamic island \cite{garcia2018ibow}. Image pairs with high similarity are deemed as possible loop closures.

Bag-of-words models, most prominently DBoW2 \cite{dbow2}, are built on the clustering of visual features. There have been  various types of feature descriptors, such as SIFT \cite{lowe2004distinctive}, SURF \cite{bay2006surf}, BRIEF \cite{calonder2010brief}, and ORB \cite{rublee2011orb}. Each of these features has its own characteristics; some are invariant towards illumination or scale but complex to compute while others may be efficient but less distinctive. These hand-crafted features are manually designed, thus none of them can be robust to all application scenarios at all times. In addition, these image representations describe the local appearance of individual patches, limiting their descriptive power with respect to global descriptor methods \cite{zhang2017loop,milford2012seqslam}.

Nevertheless, SLAM systems built on them have obtained good performance both in terms of accuracy and efficiency, and the state-of-the-art performance of ORB-SLAM \cite{mur2015orb,mur2017orb} has made itself one of the standard algorithms.

\subsection{Convolutional Neural Networks}
Convolutional neural networks are very powerful for learning visual representation by recognizing increasingly complicated visual patterns through the stacking of convolutional layers \cite{zeiler2014visualizing}. With very deep architecture design, convolutional neural networks have achieved impressive performance on classification \cite{he2016deep,huang2017densely} and object detection \cite{girshick2015fast,redmon2017yolo9000}. The ability to learn visual representations has be transferred to other tasks such as face recognition \cite{schroff2015facenet} and fine-grained classification \cite{wang2014learning}. 

The success of deep convolutional neural networks suggests its capability of learning more detailed and general representation of images. The representation can be used to accurately indicate similarity. In fact, by ranking the similarity between images in a database, deep neural networks have already been applied to image retrieval tasks \cite{gordo2016deep,radenovic2018fine}. 

There have also been some small-scale experiments applying convolutional neural networks to loop closure detection \cite{zhang2017loop,xia2017evaluation}. However, these network designs are not sufficiently utilizing the information from the environment, causing the performance to be incomparable to the state of the art from bag-of-words models. For instance, off-the-shelf usage of convolutional features did not achieve state-of-the-art performance \cite{sharif2014cnn,gordo2016deep}, unless offline data whitening is applied \cite{zhang2017loop} which is impractical in an online procedure.

Furthermore, there is a serious lack of large-scale training data adequate for training deep neural networks. In order for the networks to generalize, a dataset should contain sufficiently large numbers of images from both positive cases and negative cases. Meanwhile, there should be enough difficult loop closures that do not look very similar, as well as confusing non-closure image pairs that do look similar. However, most available loop closure datasets only contain several hundreds to thousands of images and less than 10 loop closures instances \cite{cummins2008fab,angeli2008fast}, and therefore are inadequate for training. 

Moreover, the ground truth matrices provided in many existing datasets are usually not based on the visual similarity but on scene categories (i.e., kitchen or bedroom). Other larger image datasets do not provide the ground truth for loop closures at all \cite{silberman11indoor,Silberman:ECCV12}. To the best of our knowledge, there is currently no proper dataset for the training of a deep neural network for loop closure application.

\subsection{Our Contribution} %
We address the existing problems by designing a novel Siamese architecture and train the network on large-scale datasets to obtain state-of-the-art results.

To achieve this goal and better utilize information from the environment, we add an input channel to take depth information. This provides information about the structure of the scene and is invariant to lighting conditions. The input is passed down a feature pyramid \cite{lin2017feature} to capture object representations from different scales.

We train the network end-to-end on large datasets with millions of image pairs. The datasets are simulated from Stanford 2D-3D-S Dataset \cite{armeni2017stanford} and ScanNet Dataset \cite{dai2017scannet}. We reserve one Stanford area and 3 ScanNet scenes for use as a test set. A corresponding depth image is also generated for each chromatic image as shown in Table \ref{table:dataset_examples}. We also present an algorithm for generating loop closure datasets from any similar RGB-D dataset. 

By addressing the two problems that we mentioned, our model is able to achieve state-of-the-art performance on several large-scale realistic datasets that we have labeled. Therefore, we have successfully shown for the first time that deep convolutional neural networks can be effectively applied to loop closure detection. We will release our source code and the labeled datasets to the public.

\begin{table}[h!]
\begin{center}
\begin{tabular}{|c|c|c|}
\hline
\multicolumn{3}{|c|}{Stanford} \\
\hline
Query & Positive & Negative \\
\hline
\includegraphics[width=0.29\linewidth]{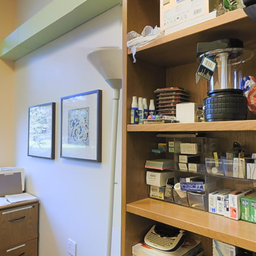} &
\includegraphics[width=0.29\linewidth]{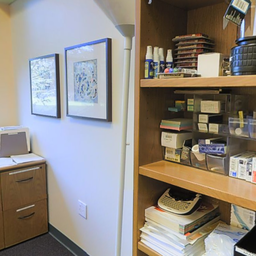} &
\includegraphics[width=0.29\linewidth]{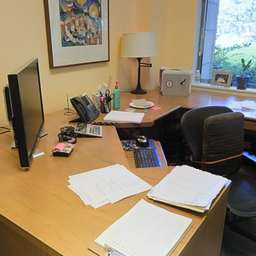}
\\
\includegraphics[width=0.29\linewidth]{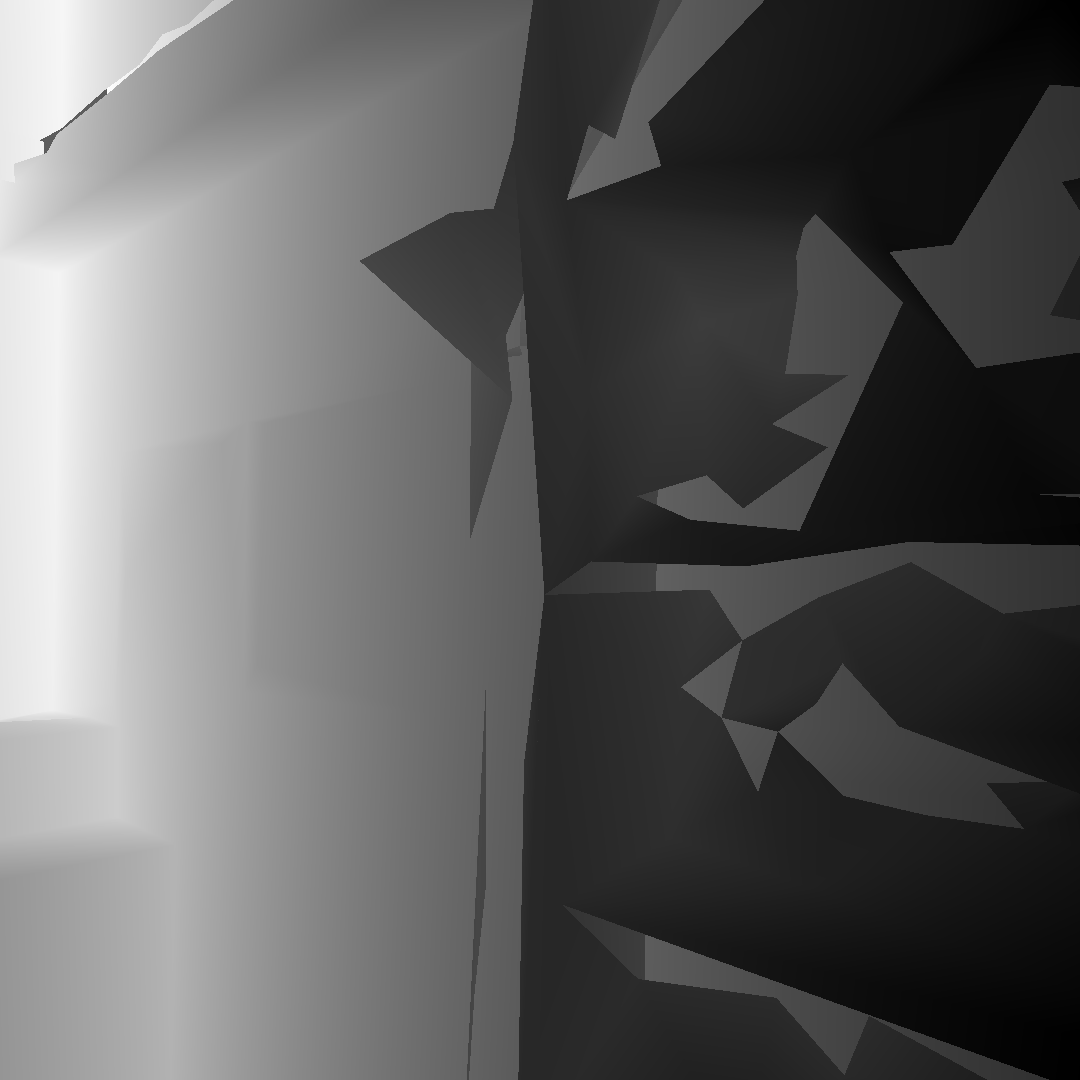} &
\includegraphics[width=0.29\linewidth]{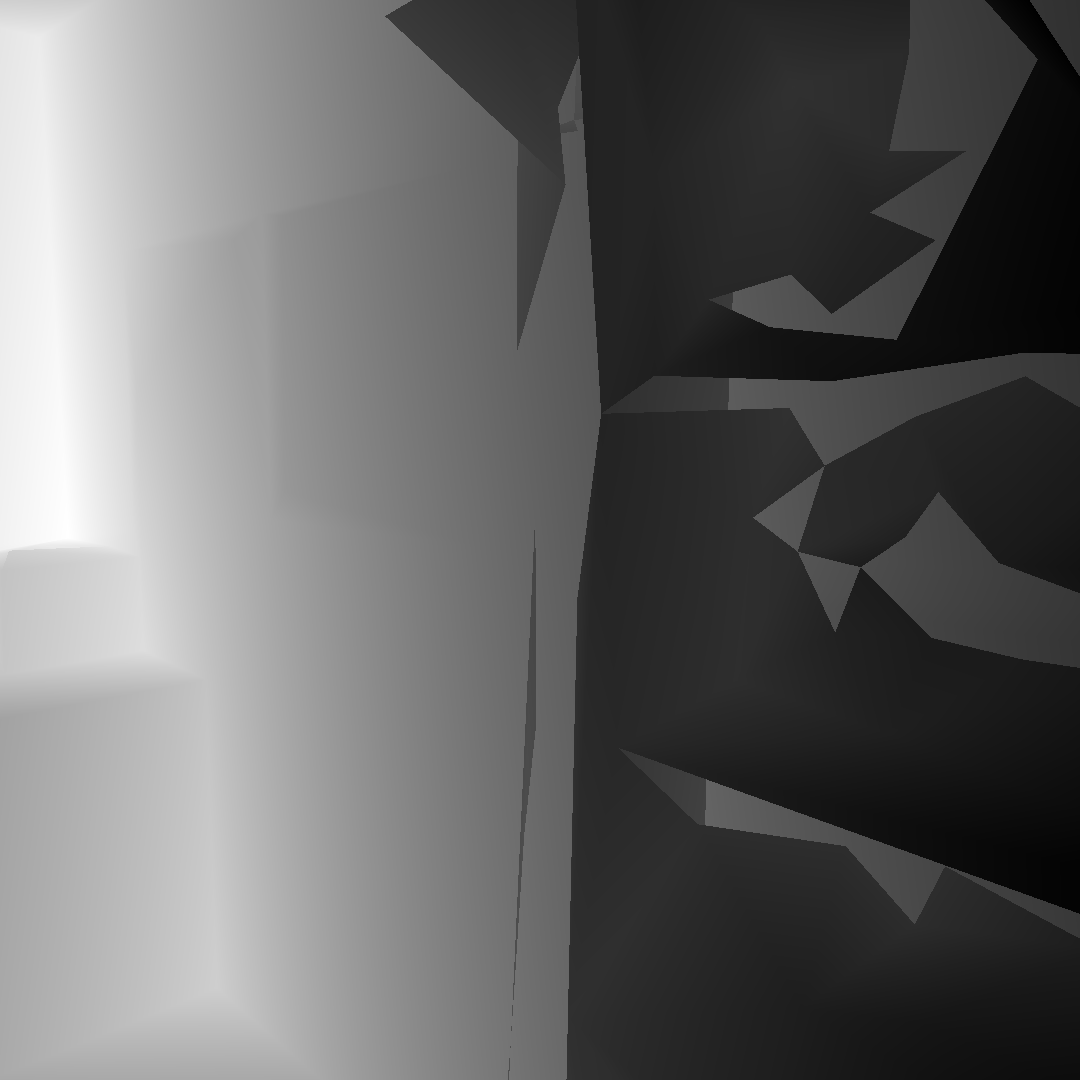} &
\includegraphics[width=0.29\linewidth]{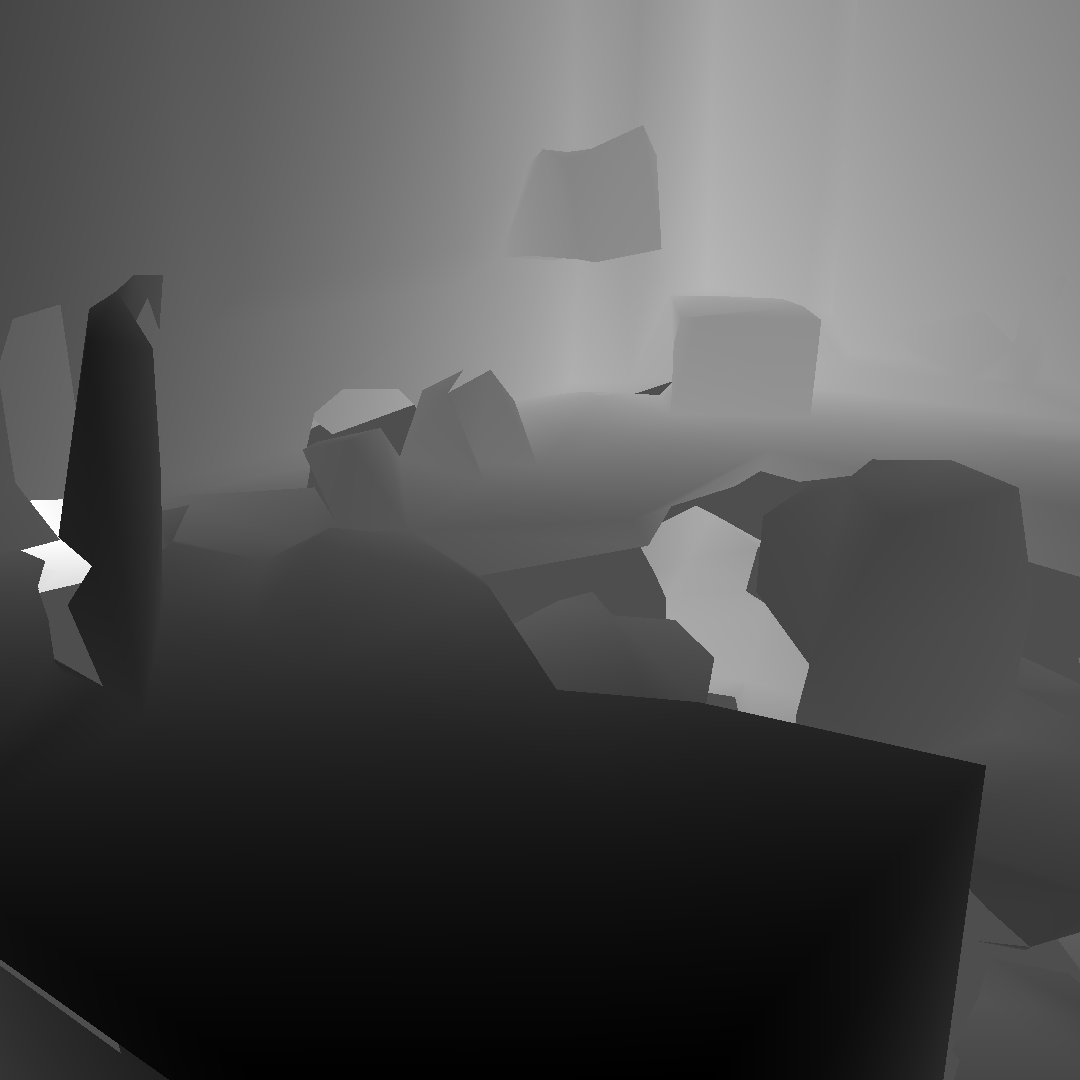}
\\
\hline
\includegraphics[width=0.29\linewidth]{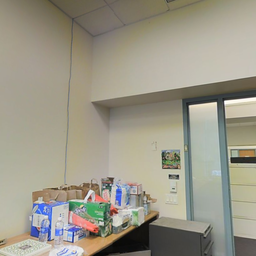} &
\includegraphics[width=0.29\linewidth]{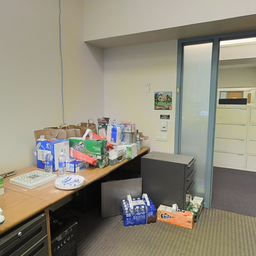} &
\includegraphics[width=0.29\linewidth]{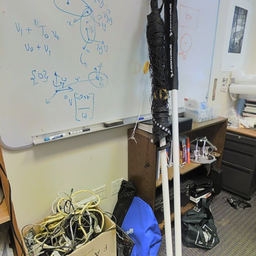}
\\
\includegraphics[width=0.29\linewidth]{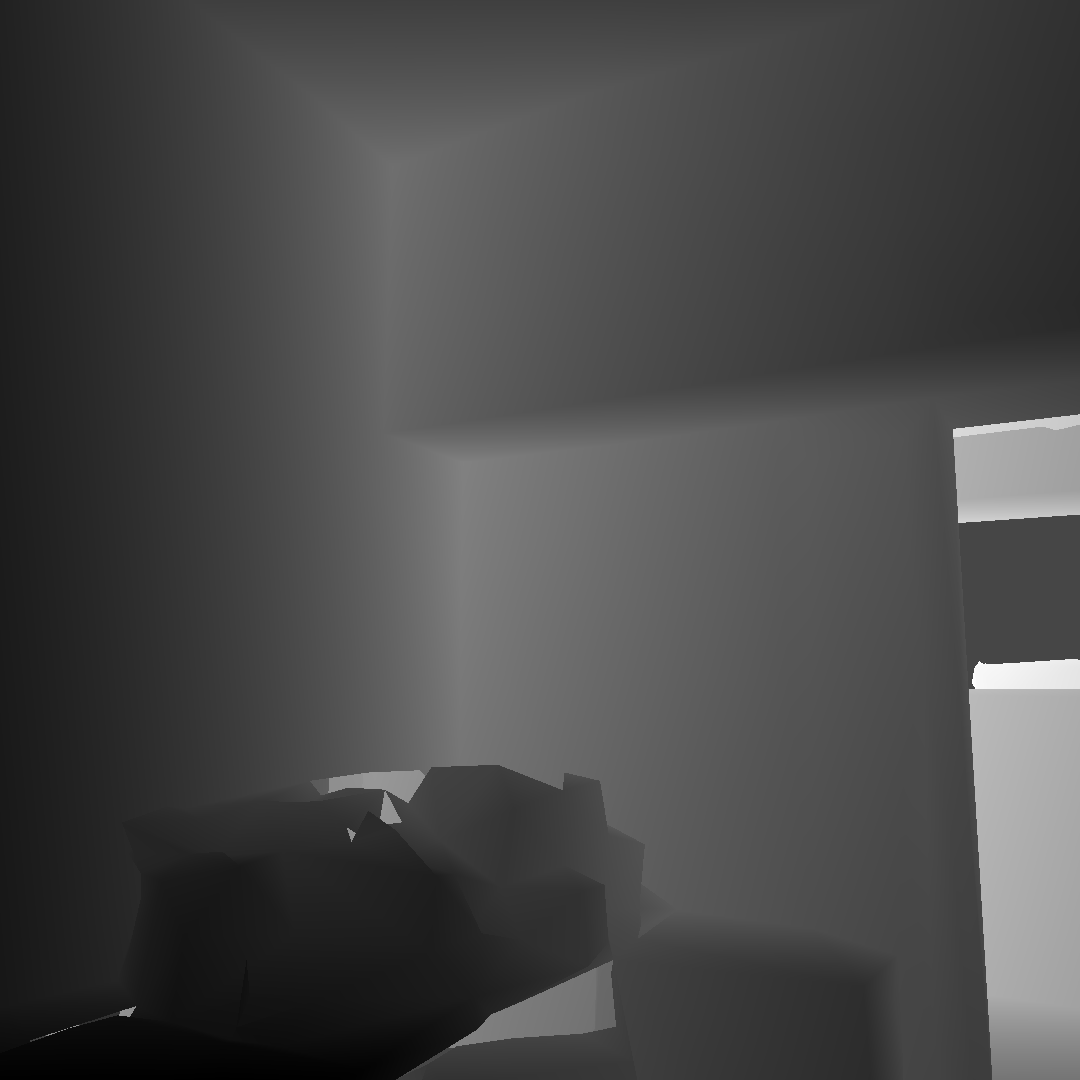} &
\includegraphics[width=0.29\linewidth]{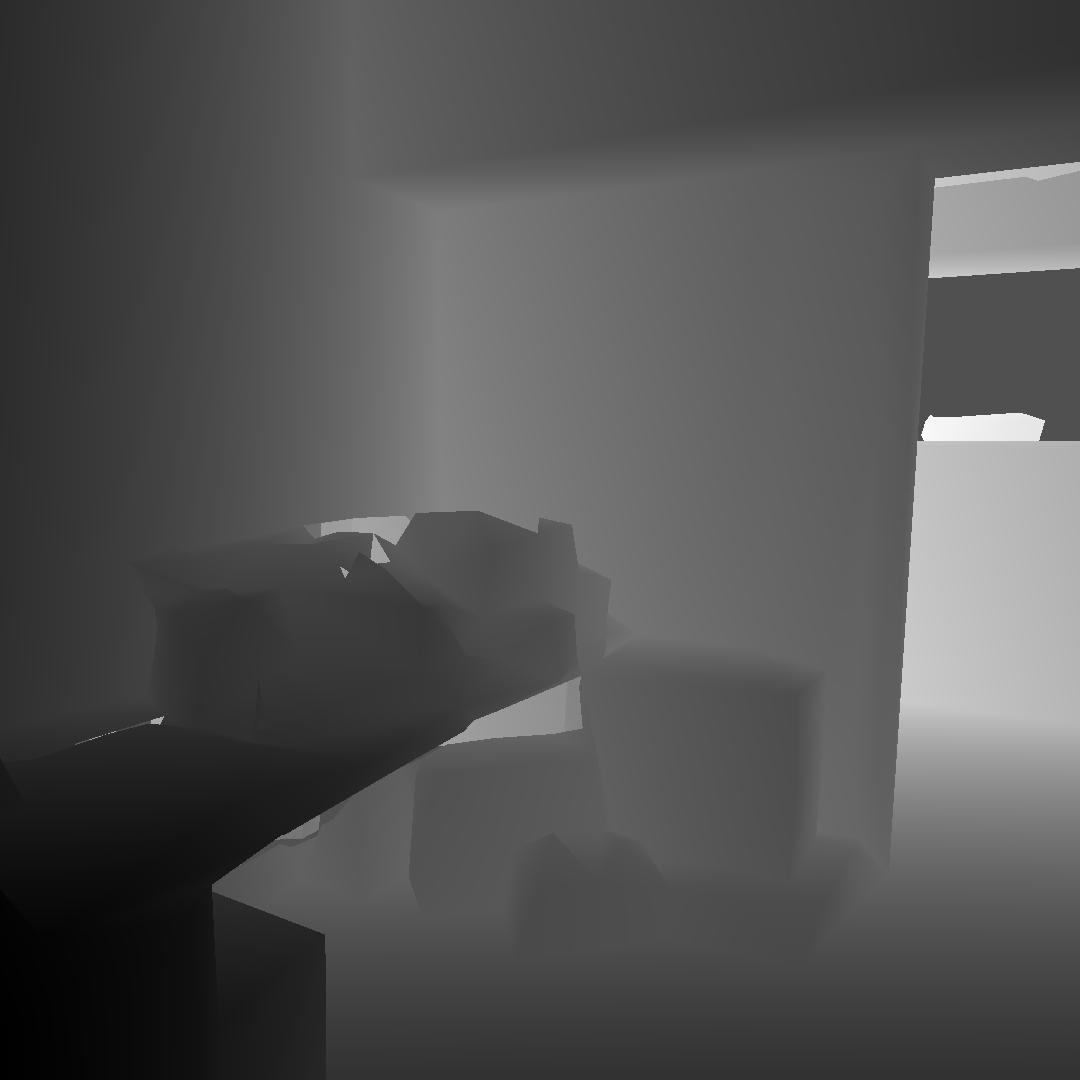} &
\includegraphics[width=0.29\linewidth]{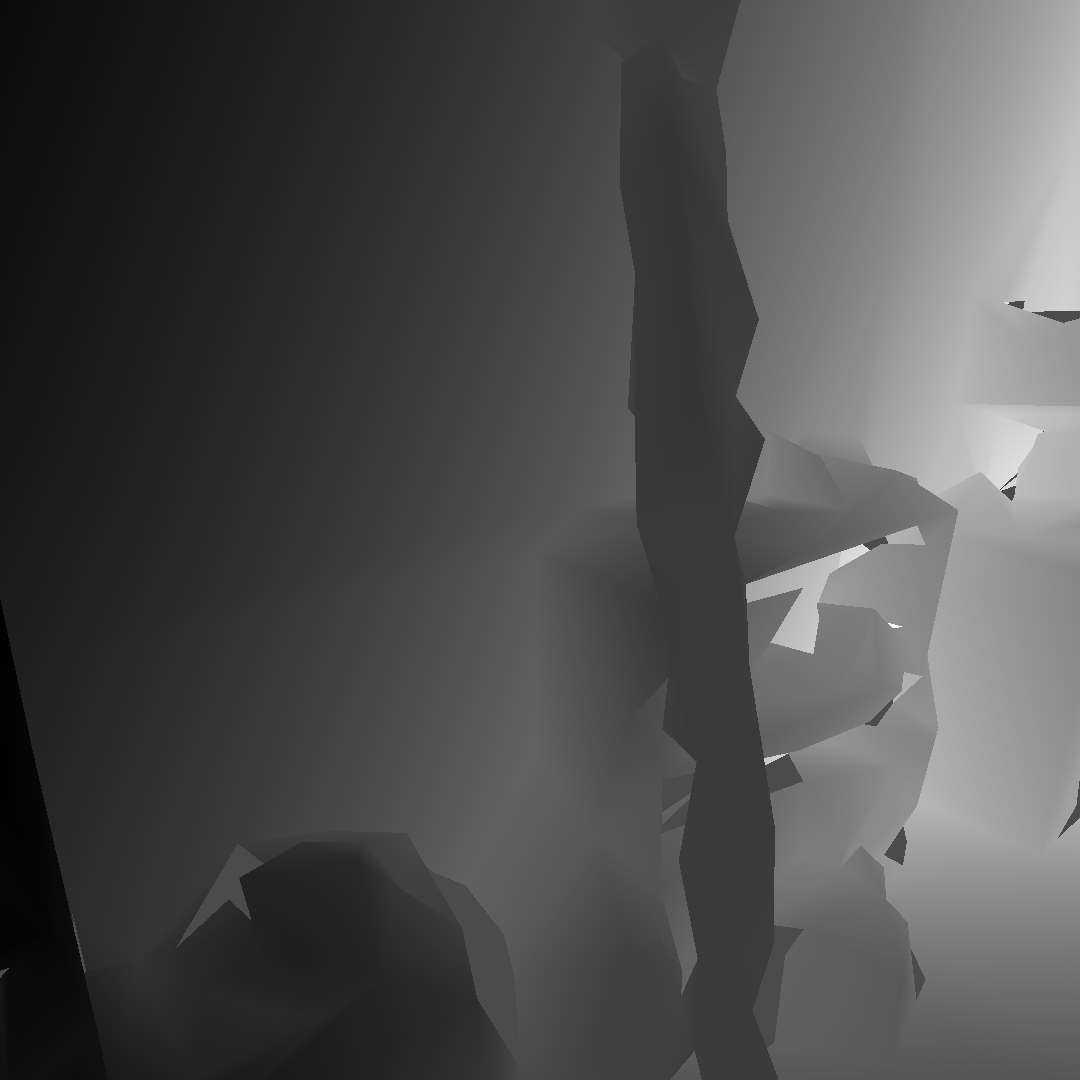}
\\
\hline\hline
\multicolumn{3}{|c|}{ScanNet} \\
\hline
Query & Positive & Negative \\
\hline
\includegraphics[width=0.29\linewidth]{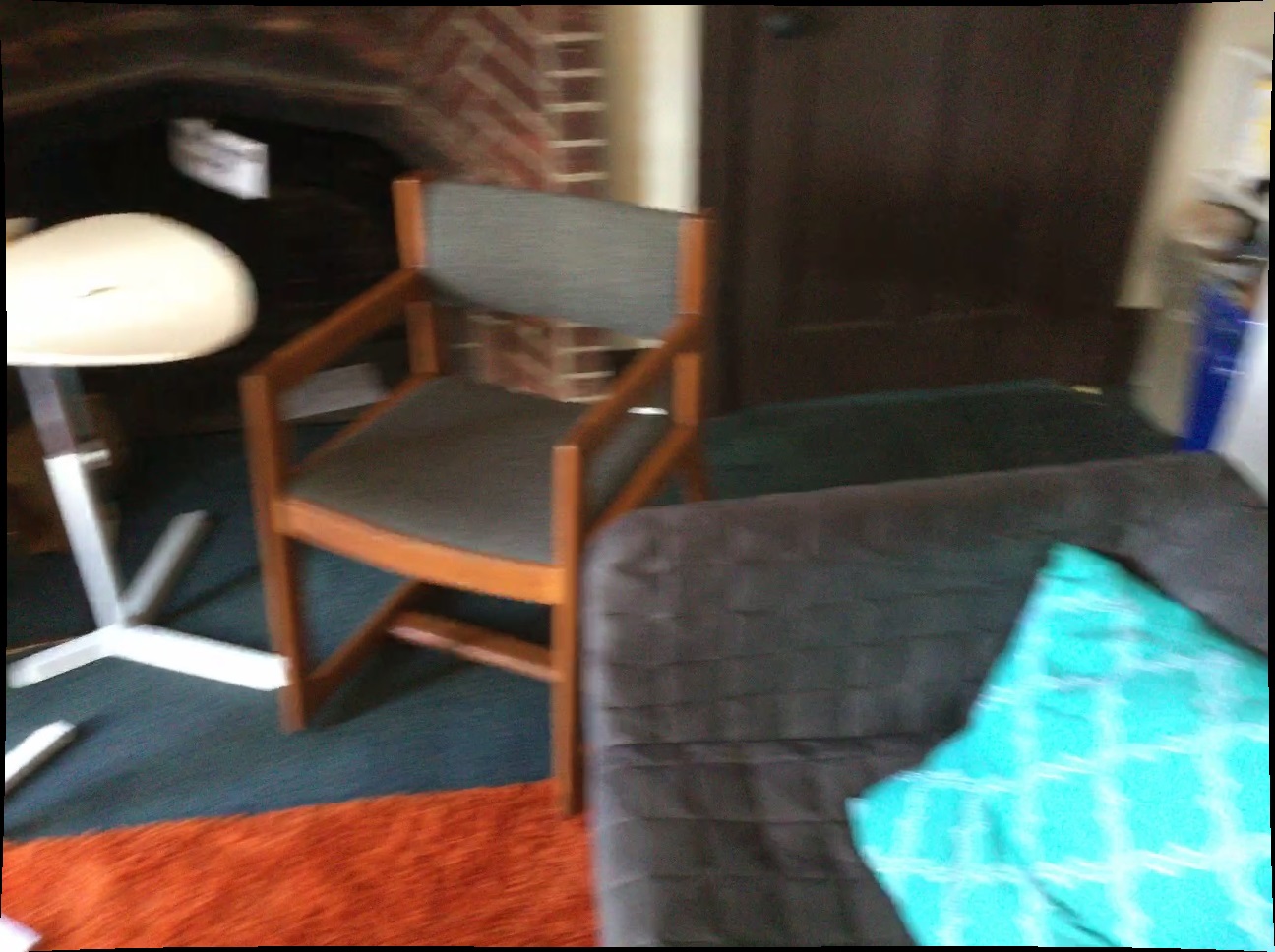} &
\includegraphics[width=0.29\linewidth]{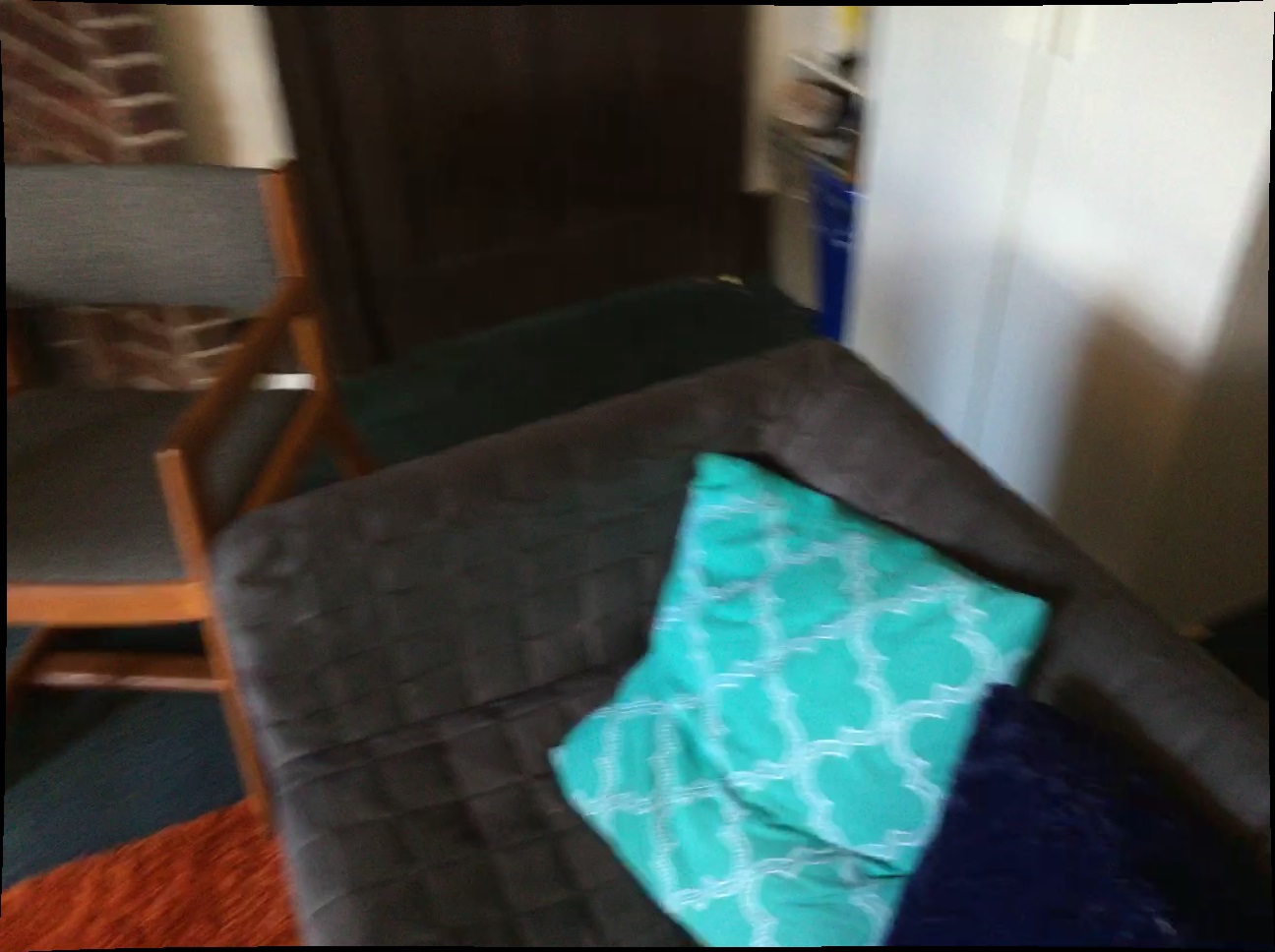} &
\includegraphics[width=0.29\linewidth]{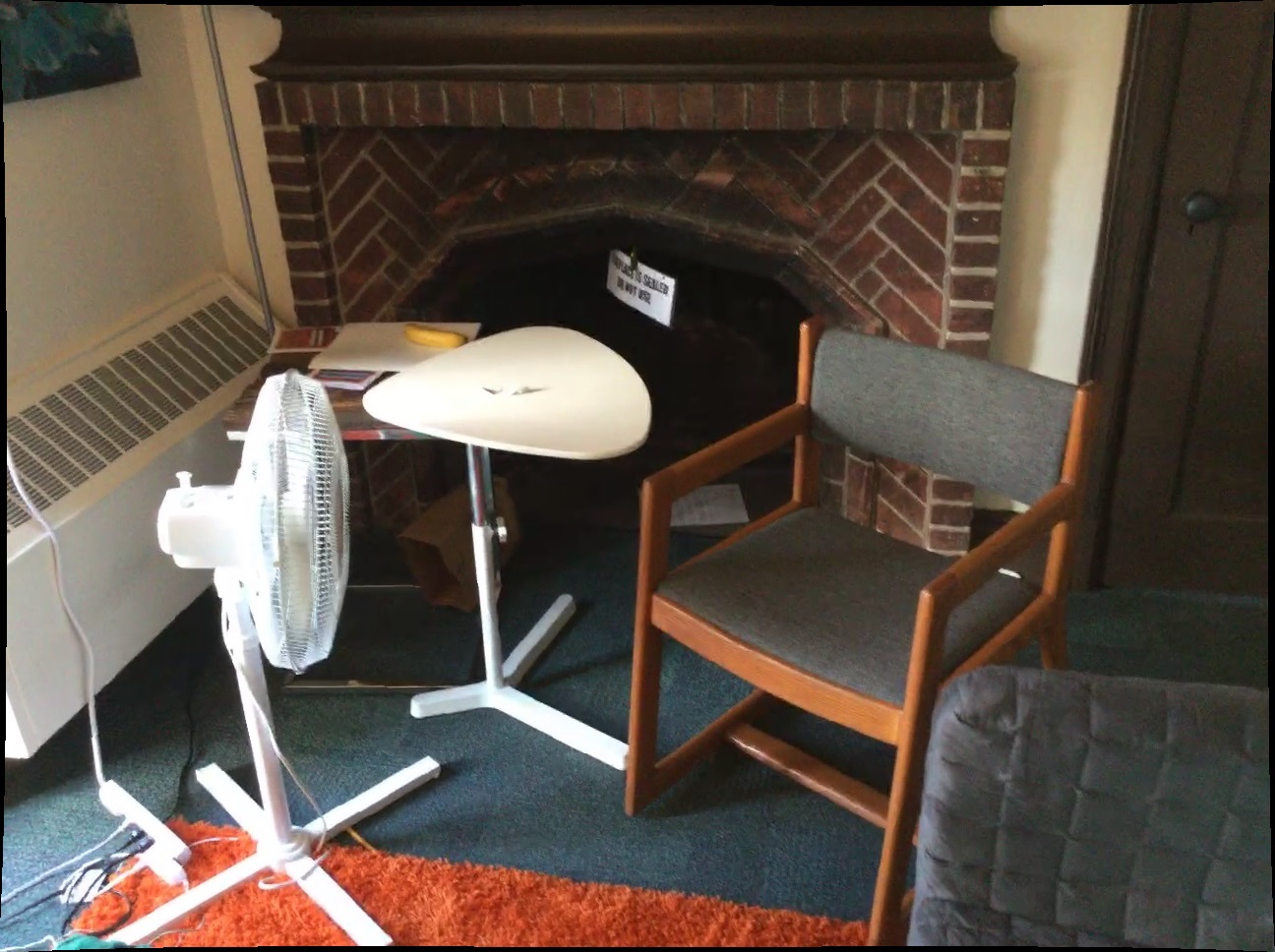}
\\
\includegraphics[width=0.29\linewidth]{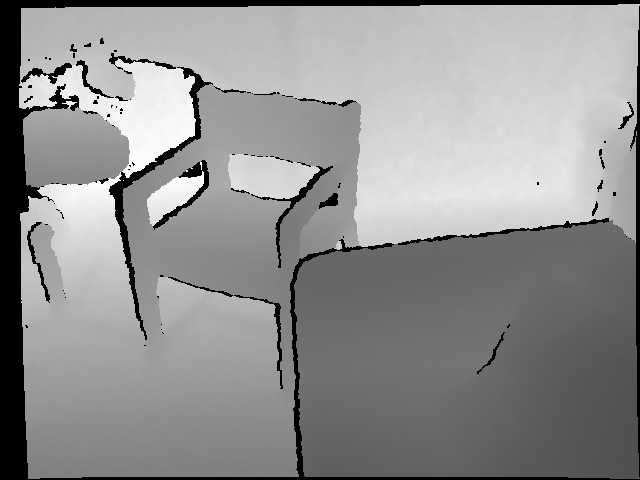} &
\includegraphics[width=0.29\linewidth]{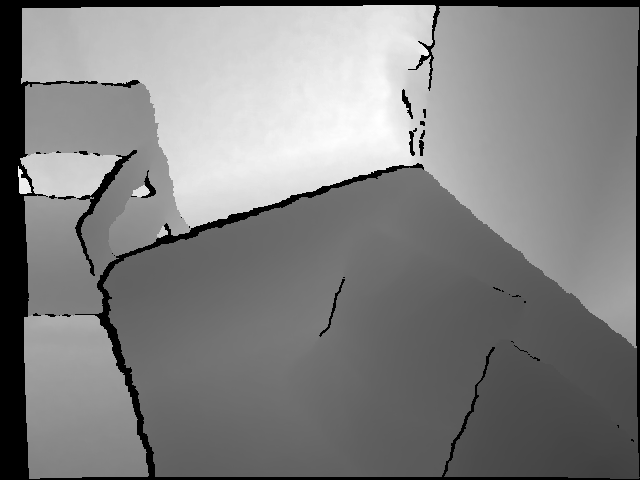} &
\includegraphics[width=0.29\linewidth]{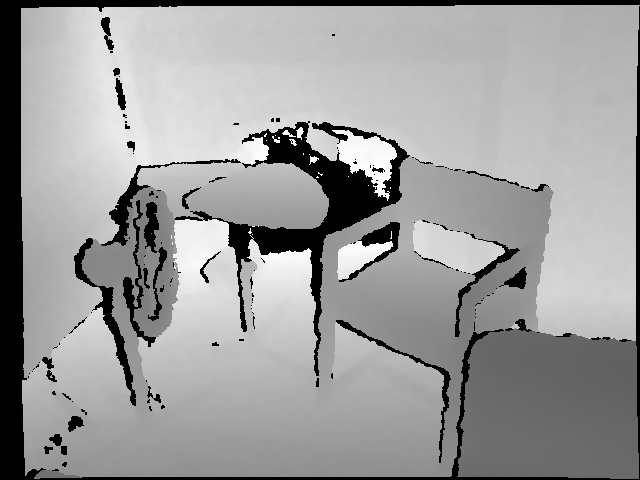}
\\
\hline
\includegraphics[width=0.29\linewidth]{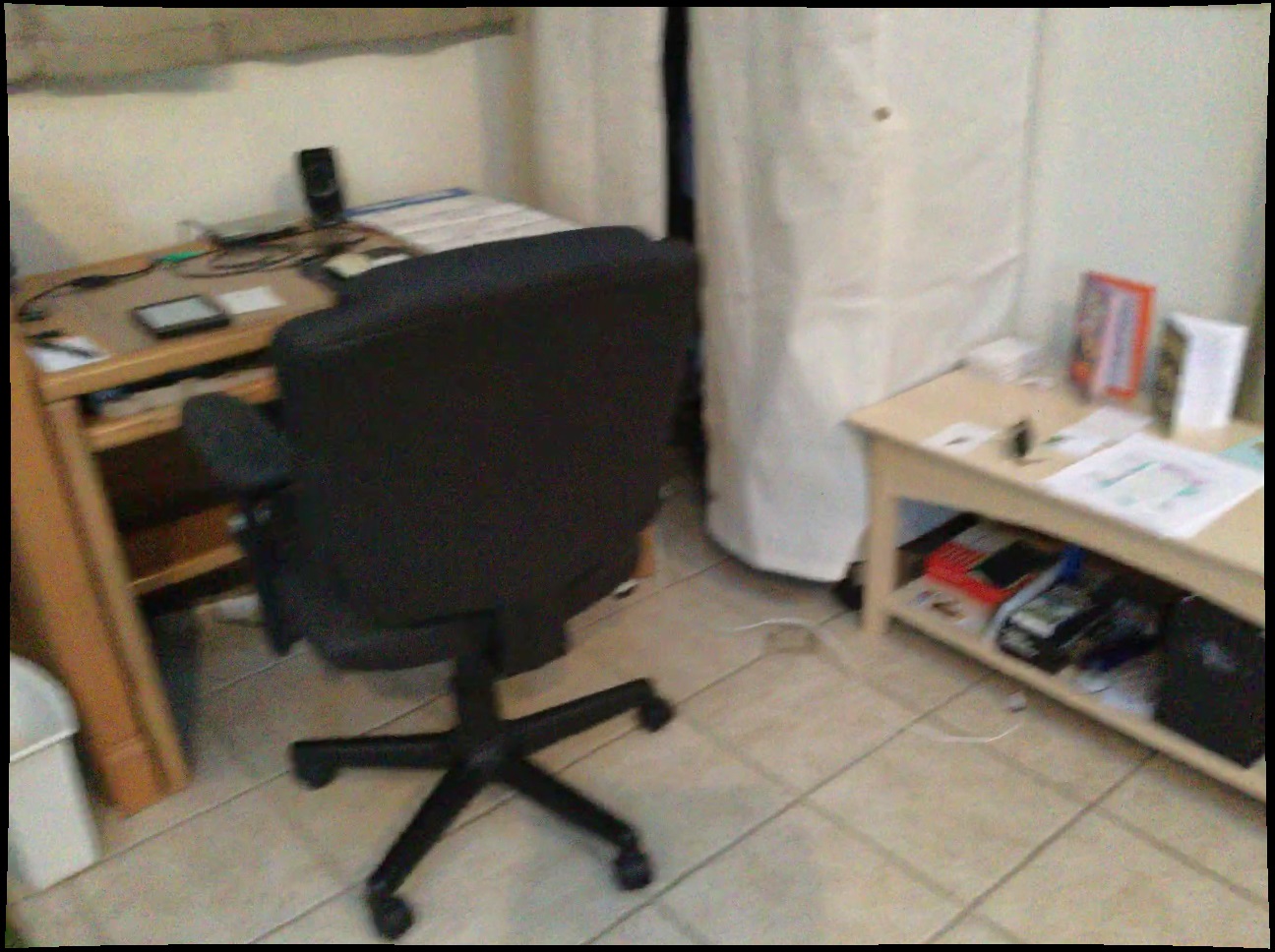} &
\includegraphics[width=0.29\linewidth]{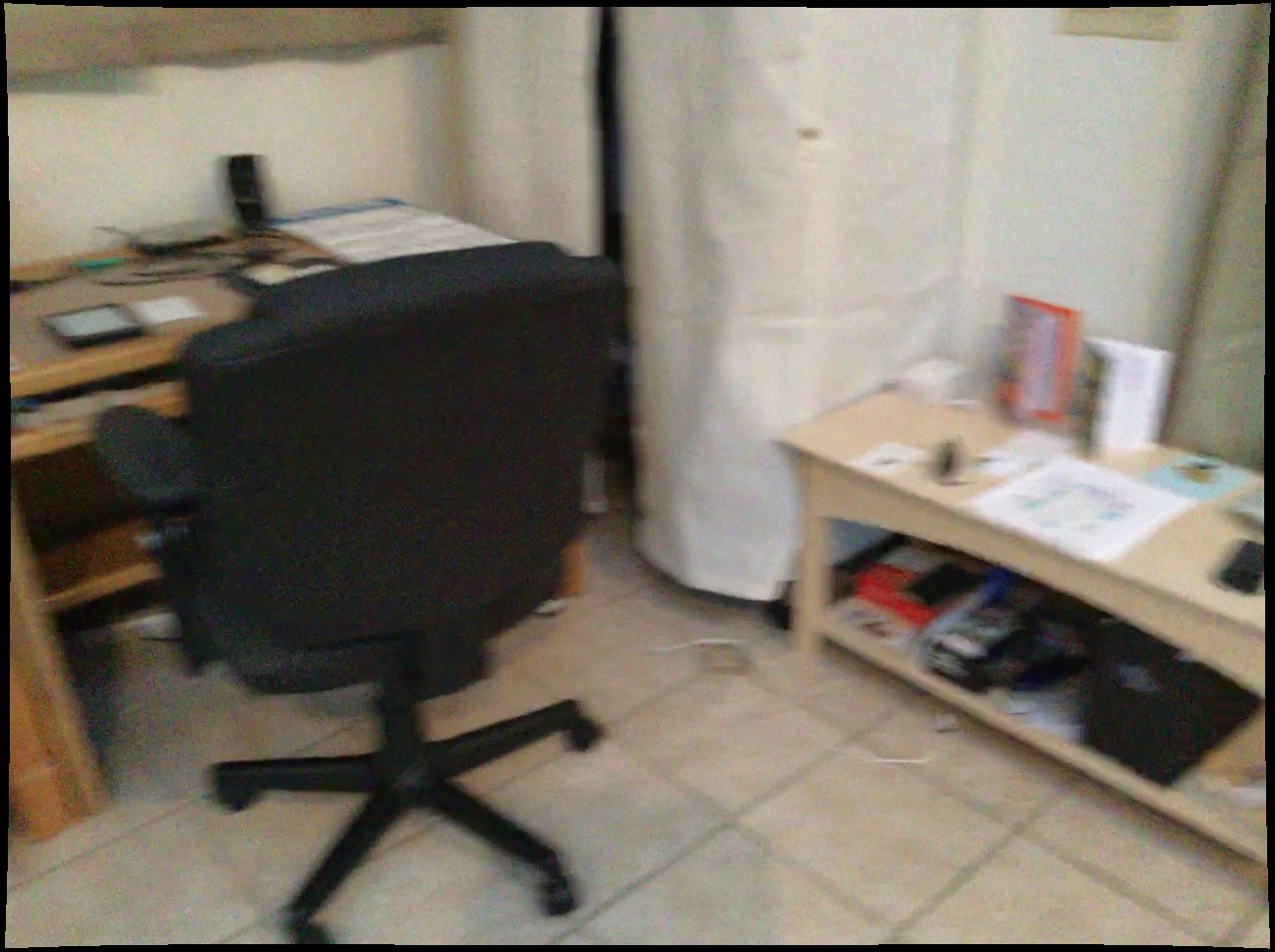} &
\includegraphics[width=0.29\linewidth]{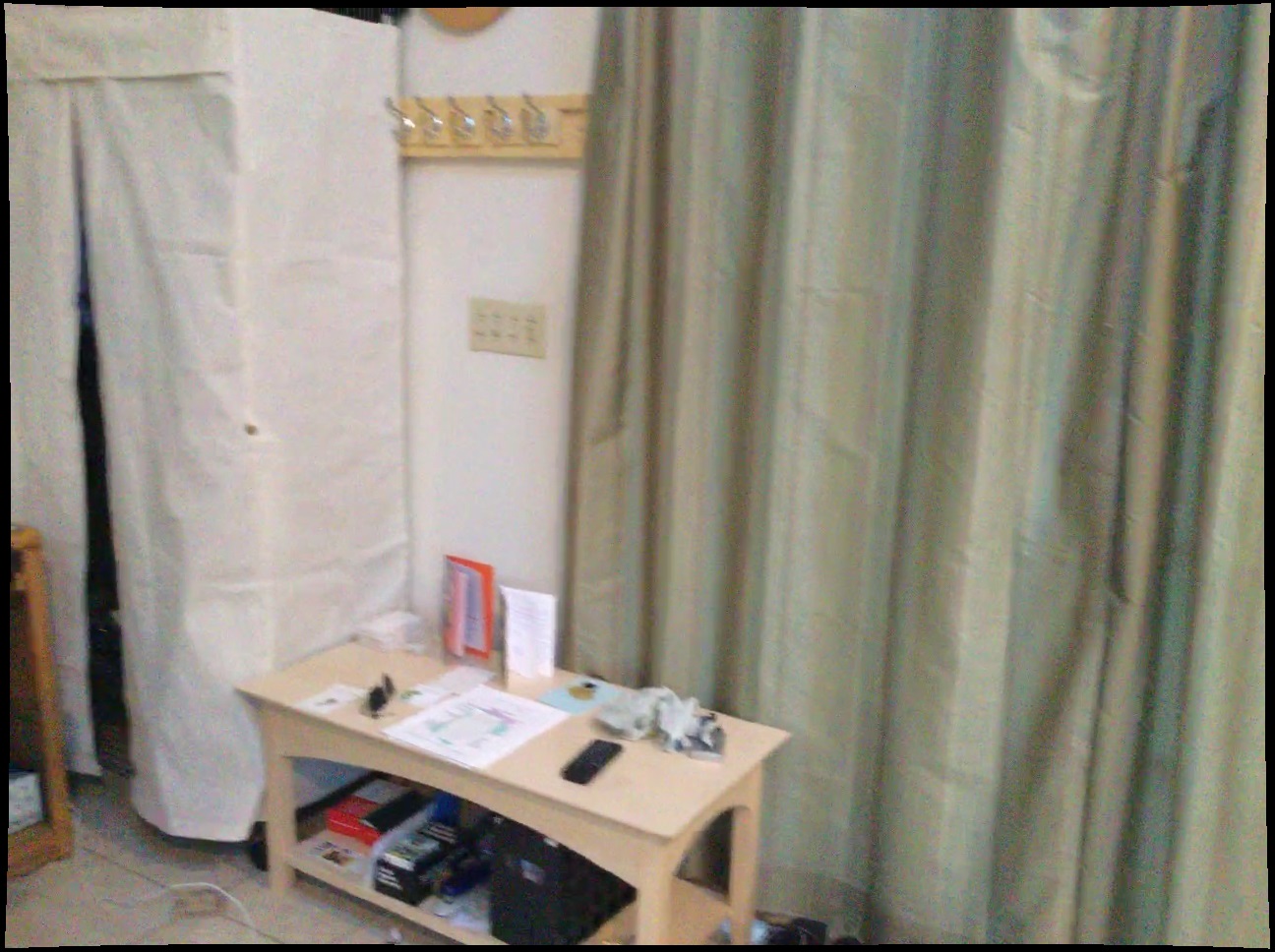}
\\ 
\includegraphics[width=0.29\linewidth]{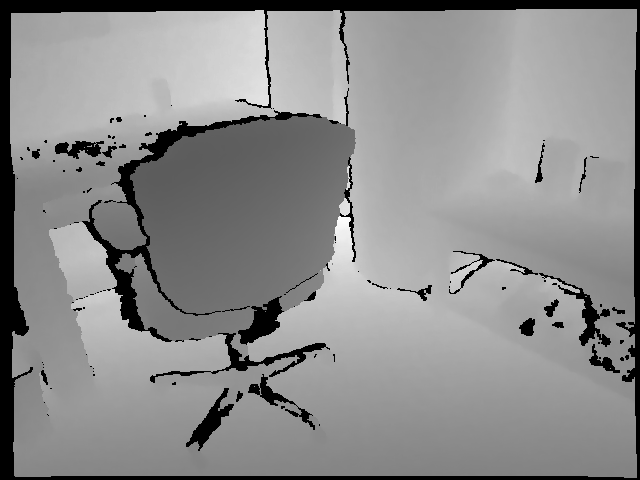} &
\includegraphics[width=0.29\linewidth]{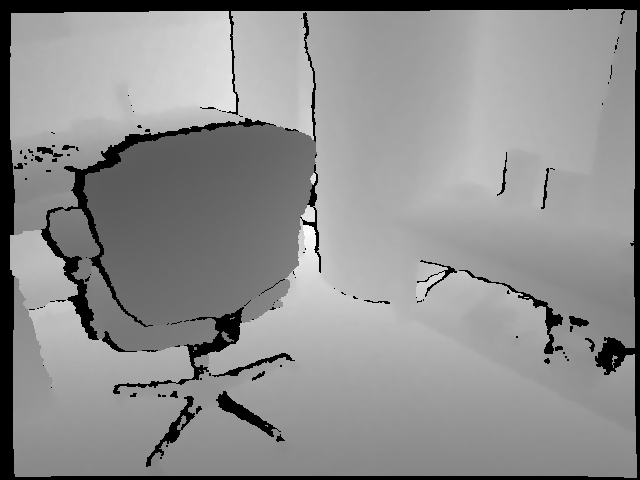} &
\includegraphics[width=0.29\linewidth]{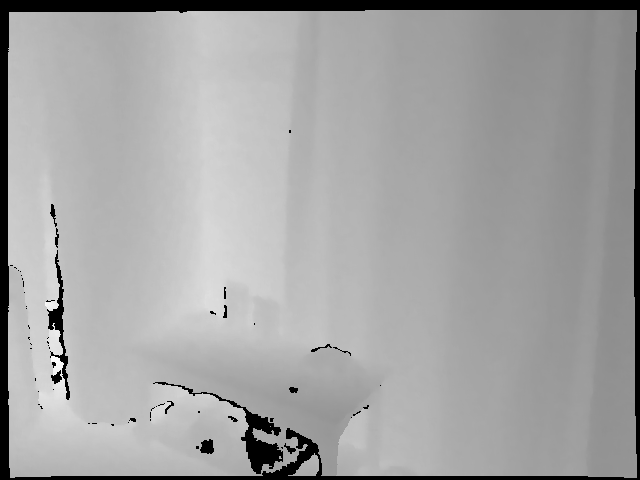}
\\
\hline
\end{tabular}
\end{center}
\caption{Examples of automatically labeled image pairs from Stanford 2D-3D-S dataset and ScanNet dataset. Depth is shown in grayscale. Query image and positive image are from the same scene, whereas negative image comes from a different scene. }
\label{table:dataset_examples}
\end{table}

\section{Proposed Network}

\begin{figure}[t]
\begin{center}
\includegraphics[width=0.8\linewidth]{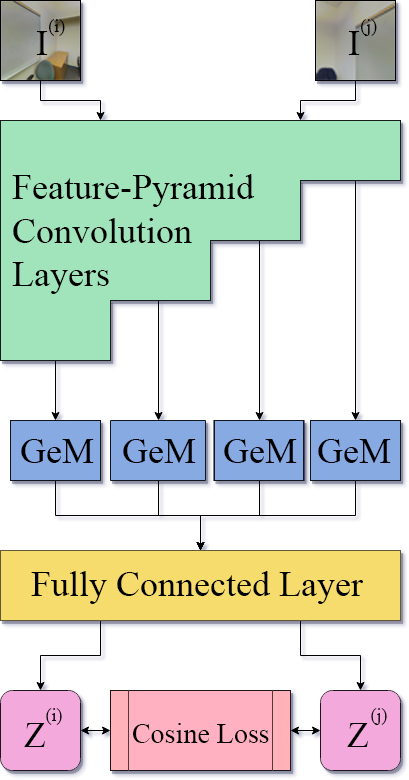}
\end{center}
   \caption{Proposed network architecture. The feature-pyramid convolution layers are extracted from ResNet50 in the four residual blocks, which then feed four feature maps to the generalized-mean pooling (GeM) layers.}
\label{fig:network}
\end{figure}

Besides off-the-shelf usage \cite{sharif2014cnn, zhang2017loop, xia2017evaluation} of convolutional features, different pooling methods have been experimented such as max pooling and sum pooling \cite{babenko2015aggregating}. Generalized-mean pooling (GeM) \cite{dollar2009integral} has provided a possibility to adjust the pooling scale between max and average with a parameter that can be learned from end-to-end training \cite{radenovic2018fine}.

The Feature Pyramid Network \cite{lin2017feature} is a novel architecture design that has achieved excellent performance on object detection. Its success in the detection of smaller objects inspires us to make use of the features from different convolution layers to accurately embed the image in different scales. 

Our network is comprised of a fully convolutional feature pyramid network $\mathcal{F}$, a set of pooling layers $\mathcal{P}_k, k \in \{1, ..., K\}$ where $K$ is the number of output scales from the feature pyramid, and a fully connected layer $\mathcal{W}$. Our proposed architecture is depicted in Figure \ref{fig:network}.

On the top, the Siamese network converts a pair of images into a pair of 2048-dimensional vectors to compute their cosine similarity. The convolutional part of the network $\mathcal{F}$ takes one image as input, and the 4 residual blocks with lateral layers output 4 feature maps in a spatial pyramid. Each feature map is fed to a different generalized-mean pooling layer $\mathcal{P}_k, k \in \{1, 2, 3, 4\}$. The 4 pooled outputs are concatenated after being passed through L2-normalization, and the concatenated feature map is whitened by a fully-connected layer $\mathcal{W}$ to generate the final output.

More formally, the network converts each pair of images into a pair of vectors $Z{(i)}$ and $Z{(j)}$. For each input image $I$ of shape $C \times H \times W$, $\mathcal{F}\left(I\right)=\{X_1, ..., X_k\}$, where $X_k$ is of shape $C_k \times W_k \times H_k, k \in \{1, ..., K\}$.
For each $X_k$, $\mathcal{P}_k\left(X_k\right) = Y_k$, where $Y_k$ is of length $C_k$.
Concatenating $\{Y_1, ..., Y_k\}$ to get $Y$ of length $\left(\sum_{k=1}^{K}C_k\right)$, $\mathcal{W}\left(Y\right)=Z$ of length $D$, and $Z$ is the final embedding of $I$.

The comparison between images is based on cosine similarity for its naturally normalized metrics and good performance on face detection \cite{wang2018cosface}. Loop closures are predicted above a certain threshold of similarity. During the training phase, the difference between ground truth and similarity provides the loss from this prediction and is used for back propagation. We refer to the proposed network as Feature Pyramid Siamese Network (FPSN).

\subsection{Feature Pyramid Convolution Layers}
\label{sec:feature-map}
The recent success of feature pyramid network on object detection tasks, especially its success in detecting smaller objects \cite{lin2017feature}, suggests that intermediate outputs of a network are inherently helpful to build semantic feature maps at different scales, which is very important to the fine-grained image comparison in loop closure detection. We thereby modify a pretrained ResNet50 into a feature pyramid model as the original paper did except that we add no padding in lateral convolution to ease the pressure on GPU-memory. The 3 intermediate outputs are passed down through sequential lateral convolution layers to generate 4 final output vectors.

Additionally, we argue that the relative distance between objects are invariant to the change of shadows and other lighting conditions, therefore depth information could be extremely useful. Noticing that the patterns of depth image is very similar to an RGB image, we copied the weight from RGB channel as the initial weight for depth channel. 

In particular, the network $\mathcal{F}$ takes an input $I$ of shape $4 \times 224 \times 224$, and outputs $X_1$ of shape $512 \times 54 \times 54$, $X_2$ of shape $512 \times 26 \times 26$, $X_3$ of shape $512 \times 12 \times 12$, and $X_4$ of shape $512 \times 5 \times 5$.

\subsection{Generalized-Mean Pooling}

The feature pyramid map generated in Section \ref{sec:feature-map} is a global descriptor, which may contain similar objects that may cause confusion during comparison. To address this problem, we add a generalized-mean pooling layer to learn to propose and pool the key regions for comparison.

Suppose the input $X$ of a generalized-mean pooling layer is of shape $C \times W \times H$, and $X_c$ is the feature map on c-th channel, then the output is given by

\begin{equation}
\begin{aligned}
\mathcal{P}\left(X\right) = \left[\mathbf{p}_1\left(X_1\right), ..., \mathbf{p}_C\left(X_C\right)\right]^T,
\end{aligned}
\end{equation}
where
\begin{equation}    
\begin{aligned} 
\mathbf{p}_c\left(X_c\right) = \left(\frac{\sum_{x \in X_c}{x^{l_c}}}{|X_c|}\right)^\frac{1}{l_c}.
\end{aligned}
\end{equation}

Average pooling and max pooling are two special examples of generalized-mean pooling. When $l_c = 1$, all elements in the feature map are accounted for equally, which makes it effectively an average pooling layer. And when $l_c \to \infty$, the pool pays it's full attention to the maximum element, which results in max pooling. The parameter $l_c$ is learned from back propagation to make an appropriate balance between these two extremes. Table \ref{table:pool_param2} and Table \ref{table:pool_param4} provide examples of different pooling parameter $l_c$ with one positive image pair taken from the first row of Table \ref{table:dataset_examples} at the second and the fourth scale of the feature pyramid. 

In particular, the pooling layers $\mathcal{P}_1, ..., \mathcal{P}_4$ outputs 4 vectors $X_1, ..., X_4$, each with length 512. The concatenated features $X$ form a 2048-dimensional vector.
\begin{table}[t!]
\begin{center}
\begin{tabular}{|c|c|c|}
\hline
$l_c = 1$ & $l_c = 3$ & $l_c = 10$\\
\hline
\includegraphics[width=0.3\linewidth]{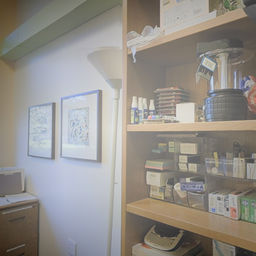} &
\includegraphics[width=0.3\linewidth]{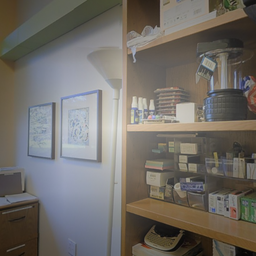} &
\includegraphics[width=0.3\linewidth]{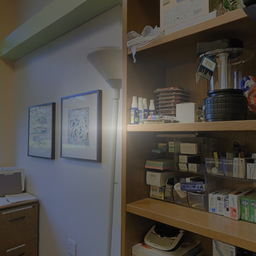}
\\
\includegraphics[width=0.3\linewidth]{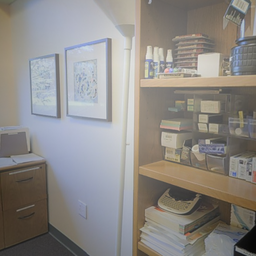} &
\includegraphics[width=0.3\linewidth]{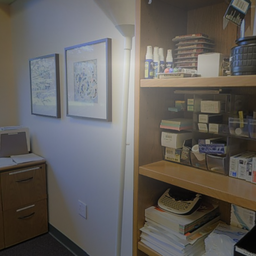} &
\includegraphics[width=0.3\linewidth]{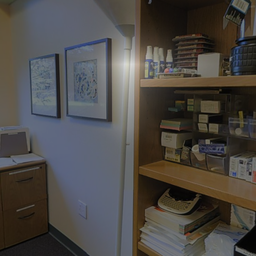}
\\
\hline
\end{tabular}
\end{center}
\caption{Visualized focus intensity of generalized-mean pooling layers with different pooling parameter $l_c$. The center of the GeM focus is on the lamp at this scale.}
\label{table:pool_param2}
\end{table}

\begin{table}[t!]
\begin{center}
\begin{tabular}{|c|c|c|}
\hline
$l_c = 1$ & $l_c = 3$ & $l_c = 10$\\
\hline
\includegraphics[width=0.3\linewidth]{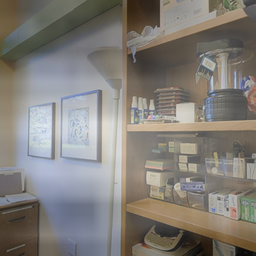} &
\includegraphics[width=0.3\linewidth]{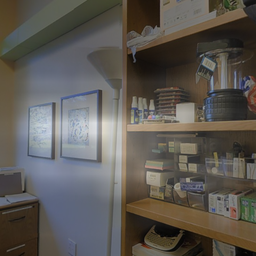} &
\includegraphics[width=0.3\linewidth]{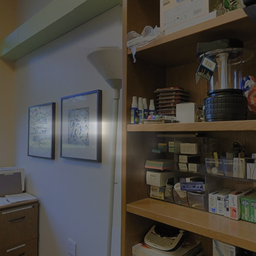}
\\
\includegraphics[width=0.3\linewidth]{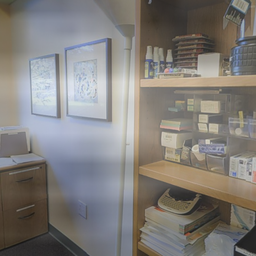} &
\includegraphics[width=0.3\linewidth]{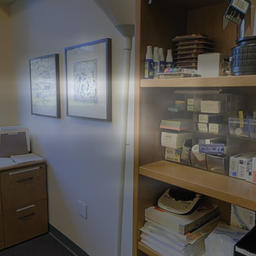} &
\includegraphics[width=0.3\linewidth]{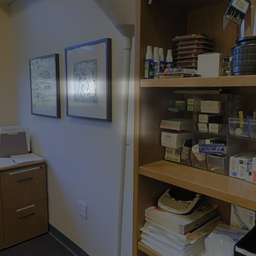}
\\
\hline
\end{tabular}
\end{center}
\caption{Visualized focus intensity of generalized-mean pooling layers with different pooling parameter $l_c$.  The center of the GeM focus is on the painting at this scale.}
\label{table:pool_param4}
\end{table}

\subsection{Fully Connected Layer}
The principle component analysis projection, in its mathematical form, is equivalent to a fully connected layer. Therefore, to make the network entirely end-to-end, we add a fully connected layer to perform online data whitening \cite{gordo2016deep} instead of principle component analysis offline \cite{zhang2017loop}.

We maintain the dimensionality to preserve the information while performing discriminative large-margin metric in which one learns a new space where relevant images are closer. The layer takes 2048-dimensional vector and outputs a vector of the same length as final encoding of the input image from the network.

\subsection{Cosine Loss}
The output vector is compared in pairs. The model takes two images $I^{(i)}$ and $I^{(j)}$, and compute the similarity between their embedding $Z^{(i)}$ and $Z^{(j)}$. We use cosine similarity, defined as: 
\begin{equation}    
\begin{aligned} 
\mbox{similarity}(Z^{(i)}, Z^{(j)}) = \frac{Z^{(i)} \cdot Z^{(j)}}{\|Z^{(i)}\| \times \|Z^{(j)}\|}.
\end{aligned}
\end{equation}

The ground truth of pairs is 1 if it is a loop closure, 2 for unusable pairs, or 0 otherwise. The loss is computed based on the difference between predicted similarity and ground truth. A margin is set to further distinguish positive cases from negative cases. 

Let $g$ be the ground truth, $s$ be the predicted cosine similarity, $m$ be the margin. The formula for the loss is 
\begin{equation}    
\begin{aligned} 
\mbox{loss}(s) = \begin{cases}
          1 - s, & \text{if $g=1$;} \\
          \mbox{max}(0, s - m), & \text{if $g=0$.}
          \end{cases}
\end{aligned}
\end{equation}


\section{Generating Large-Scale Datasets}
\subsection{Traditional Test Sets are Insufficient}
In the literature, New College dataset \cite{cummins2008fab}, City Centre dataset \cite{cummins2008fab}, Lip6 Indoor dataset \cite{angeli2008fast}, and Lip6 Outdoor dataset \cite{angeli2008fast} are some of the most used datasets for loop closure detection. Although these datasets can have several hundred loop closure pairs, typically they only have one loop closure sequence, which provides almost no variety in the positive pairs tested. In addition, often the way these loop closure datasets is labeled is often not well suited for pairwise comparison. In fact, Lip6Indoor and Lip6Outdoor have asymmetric ground truth matrices and do not mark images along the diagonal as positives.

We developed an algorithm to generate a larger and more diverse loop closure dataset by detecting loop closures in an offline manner with much more information than what would be available to an agent to address this problem. There are very large datasets that have quality depth and pose estimation, which we utilize to generate training and testing data for our neural network. The algorithm is also open source so other people may develop their own datasets and refine their loop closure algorithms.

\subsection{Stanford 2D-3D-S Dataset}
The Stanford 2D-3D-S dataset  \cite{armeni2017stanford} contains 70,496 RGB-D images that originate from 3 different buildings of mainly educational and office use. The dataset is collected in 6 large-scale indoor areas covering over 6,000 $m^2$ using the Matterport Camera, which combines 3 structured-light sensors at different pitches to capture 18 RGB and depth images during a 360° rotation at each scan location. Each 360° sweep is performed in increments of 60°, providing 6 triplets of RGB-D data per location. The output is the reconstructed 3D textured meshes of the scanned area, the raw RGB-D images, and camera metadata. This data is then post processed to refine the depth of each image in conjunction with it's pose. 

\begin{figure}
\begin{center}
\includegraphics[width=1.0\linewidth]{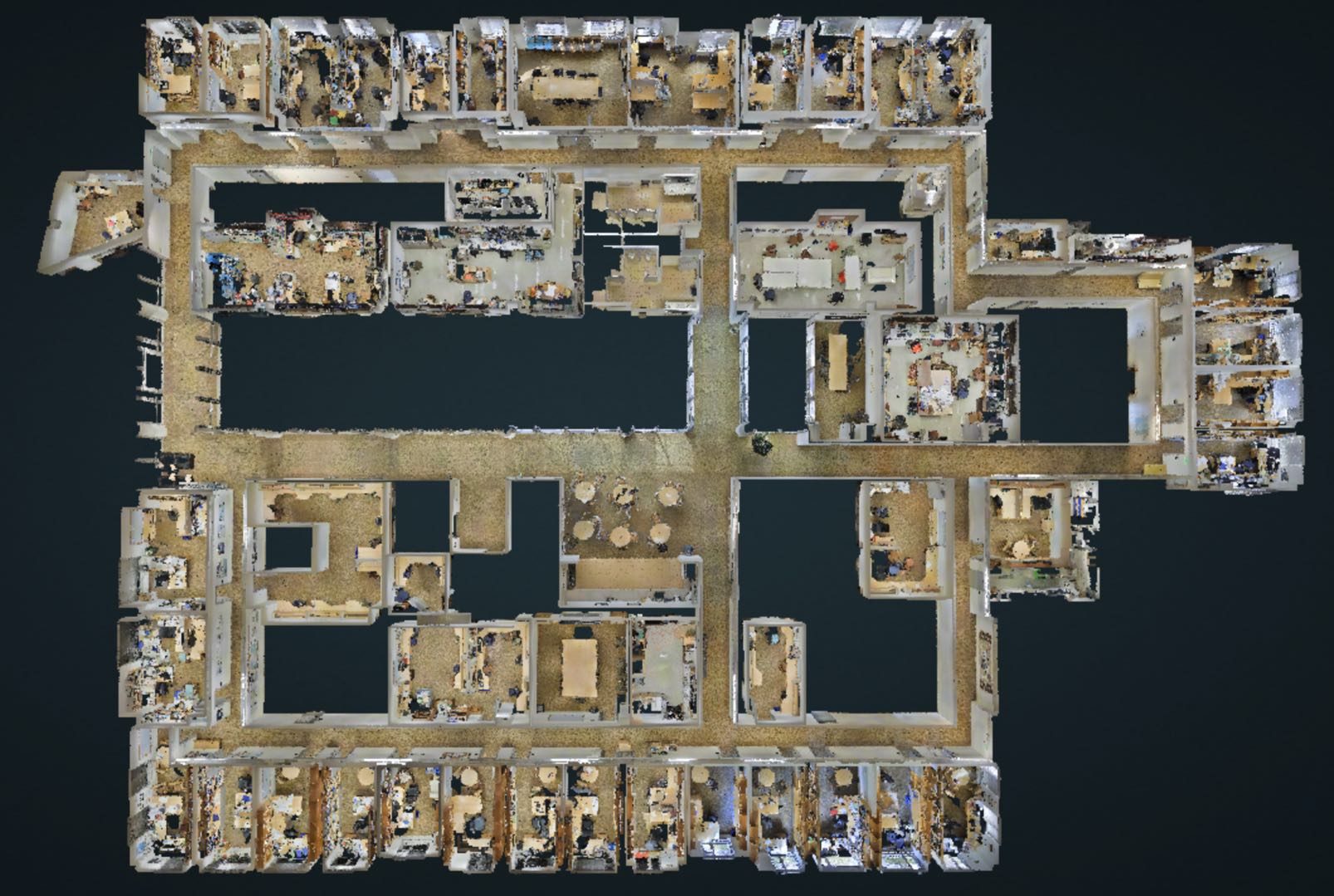}
\end{center}
   \caption{Overview of the Stanford dataset area from which the testing dataset is sampled \cite{armeni2017stanford}.}
\label{fig:area5}
\end{figure}

\begin{figure}
\begin{center}
\includegraphics[width=1.0\linewidth]{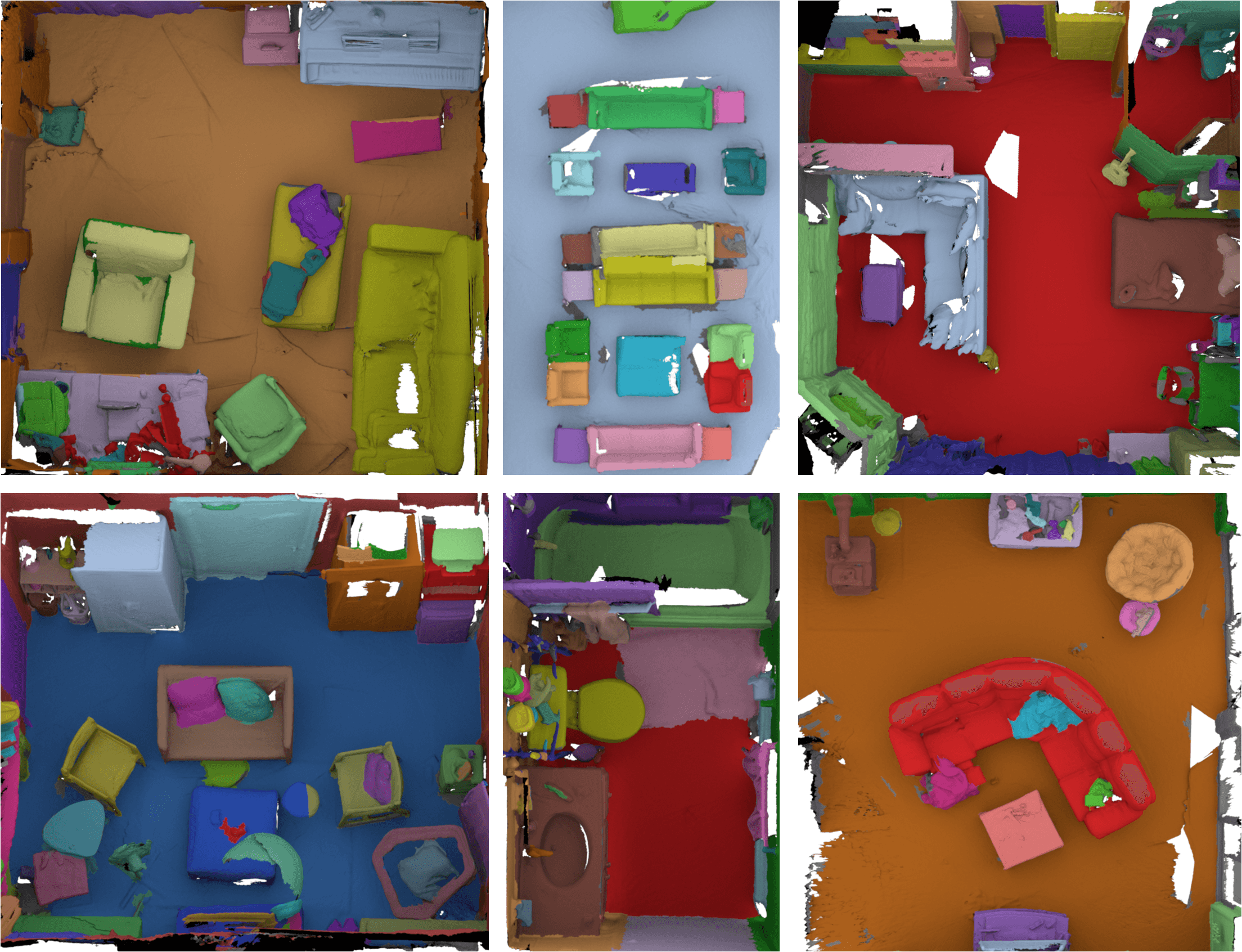}
\end{center}
   \caption{Overview of scenes from ScanNet. The objects are color coded for annotation \cite{dai2017scannet}.}
\label{fig:scannet}
\end{figure}

\subsection{ScanNet Dataset}
ScanNet \cite{dai2017scannet} is an RGB-D video dataset containing 2.5 million views in 1,513 RGB-D scans of 707 unique indoor environments collected using the Occipital Structure RGB-D sensor \cite{occipital2016structure}. The Occipital sensor collects 640x480 images at 30 Hz, similar to the Microsoft Kinect. ScanNet contains a variety of small spaces such as offices, apartments, and bathrooms. Each scan has been annotated with instance-level semantic category labels through crowd sourcing. We select the largest scans from this dataset for testing as many of the rooms are too small for use in loop closure.

\subsection{Automatic Loop Closure Labeling}
Each image pair from an RGB-D dataset is assigned a score based on backprojecting the point cloud from one image into the other, which we use to separate positive from negative pairs. However, there are a few more steps to prevent false positives and speed up the process.

We first subsample the dataset based on a fixed ratio to avoid generating too many pairs of similar images. We also filter out images that have too little texture or do not have valid depth information. 

The volumetric overlap between the two point clouds is then calculated by comparing the convex hull from the pair of point clouds. The volume of the intersection between the two hulls is divided by the volume of the larger of the two hulls is used to filter out pairs of images before moving on to the next step. Images with low overlap are marked as low confidence negatives and we do not use them during training or testing.

For the final step, for each pair we backproject the point cloud for one image into the the camera of the other, then downsample and threshold the result to obtain the percent coverage the second image represents in the first. We project the location of each point of the point cloud into the coordinates in the image space provided that the depth camera is fully calibrated. We then downsample the image to compensate for the sparseness of the point cloud then count the number of non-zero pixels. The percent image overlap is the confidence associated with each image pair. For our purposes, we mark images with greater than 50\% as a positive pair and don't use the rest. 

From Stanford 2D-3D-S dataset, we automatically labeled around 25,434 images from 6 areas: area1, area3, area4, area5a, area5b, area6. We reserve area 5a with 5,000 images for testing purposes, and use all the other areas for training. The training set consists of around 21,000 images that yield millions of usable image pairs. On average, image pairs that are labelled as loop closures take up around 1 in every 400 usable pairs, which we believe is consistent with the probability of loop closure occurring in large-scale indoor navigation scenarios. 

From ScanNet dataset, we similarly label images from 3 scenes: scene0000\_01, scene0000\_02 and scene0002\_01. Because the data are captured by relatively more inexpensive device, the motion blur in the RGB image and errors in depth are more severe compared to those from Stanford 2D-3D-S. All three of these sample sets are used for testing purposes.

\section{Experiment}
\subsection{Training Procedure}
We start from training a plain ResNet50 \cite{he2016deep} embedder as baseline. The architecture is identical to a ResNet50 except discarding the last pooling layer and the fully-connected layer at the end. We instead feed the output of the last convolutional layer to a generalized-mean pooling layer, and then to a fully connected layer after L2-normalization. 

The weights of convolutional layers are initialized from the pre-trained classification model on ImageNet. The initial pooling parameter is 3 which turns out to be very close to the training result, and the weight of fully-connected layer is initialized by random Gaussian distribution.

We use 4 GeForce GTX 1080 Ti GPUs for training. Multiple images are associated to each query to reduce GPU-memory consumption. Specifically, either 7 or 352 images are loaded in a tuple for every query. In each tuple, the first image is always the query, and the second is always the positive image; all the other images in this tuple are negative. 

The training iterates two stages. In one stage we use 32 tuples of 7 images to learn the similarity between query and positive images. In the other stage we use 1 tuple that contains 352 images, to bring the ratio between positive and negative cases (1:350), which is as close to the ratio found in the training dataset as can be obtained within the constraints of GPU-memory. At this stage the model tries to distinguish real loop closures from similar but different image pairs. 

To obtain the benefits of both Adam for fast convergence and Stochastic Gradient Descent (SGD) for better generalization \cite{nitish2017improving}, we start the first round of 2 stages with Adam with \textit{1e-6} learning rate, \textit{0.9} momentum and \textit{5e-5} weight decay for 24 hours each stage. Then we switch to SGD with \textit{1e-5} learning rate, \textit{0.9} momentum and \textit{5e-5} weight decay. After 3 iterations, the model converges to over \textit{99.9\%} accuracy on the training set.

Then, we train a 4-channel ResNet50 embedder. The procedure is identical to the that of a plain ResNet50 above except that the first convolutional layer takes input in 4 channels, where the initial weight of the fourth channel is copied from the third channel. Also we adjust the limit for number of images to 302 in the second stage to account for the change in memory capacity. Finally, we train the feature pyramid model initialized from the 4-channel ResNet50 model, in which the image number at stage 2 is limited to 252.

\subsection{Testing Procedure} 
We reserve area 5a from the Stanford 2D-3D-S dataset and scene0000\_01, scene0000\_02 and scene0002\_01 from ScanNet for testing. Some examples can be seen in Table \ref{table:dataset_examples}.

Area 5a is a typical teaching building at Stanford University. The dataset for this area contains 5,000 $256 \times 256$ RGB and associated depth images of the same dimensions. The entire RGB-D image is then resized to be of size $224 \times 224$ per channel. Each depth image is scaled down within [0, 1] by min-max normalization. Any pixel with a depth beyond 6 meters in the image indicates that the pixel is unusable, so we set it's depth value to 0. In total, there are 8,743,938 negative pairs and 25,036 positive pairs in the dataset in a ratio of roughly 350:1.

The same preprocessing of depth information is applied to ScanNet scenes. Scene0000\_01 contains 197 images with 23,576 negative pairs and 331 positive pairs (71:1). Scene0000\_02 contains 102 images with 6,264 negative pairs and 152 positive pairs (41:1). Scene0002\_01 contains 241 images with 31,420 negative pairs and 259 positive pairs (121:1).

The model predictions for each dataset are stored as a matrix $S$, where each element $S_{ij}$ holds the cosine similarity between image $i$ and image $j$. The matrix is compared directly against the ground truth matrix $G$, as we see if $S_{ij}$ is above certain threshold if $G_{ij}$ is 1 (positive pair), and below that threshold if $G_{ij}$ is 0 (negative pair). We thereby compute the number of true positive, true negative, false positive and false negative cases for each value of the threshold. By altering the threshold from 0 to 1, we compute the corresponding precision and recall, and then plot the precision-recall curve for the dataset.

\subsection{Results}
\begin{figure*}
\begin{center}
\begin{tabular}{cc}
\includegraphics[width=0.4\linewidth]{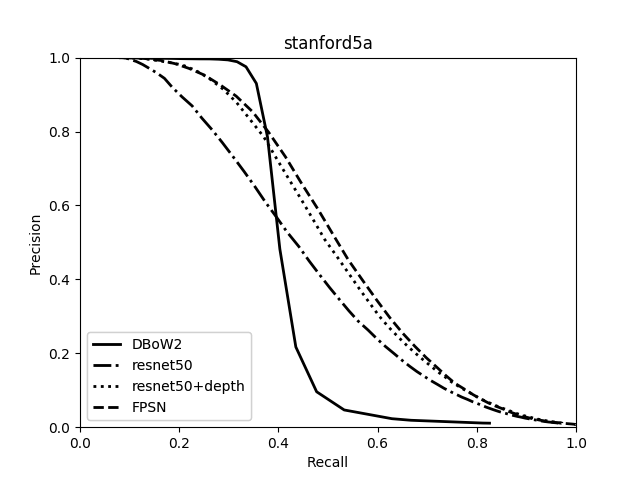} &
\includegraphics[width=0.4\linewidth]{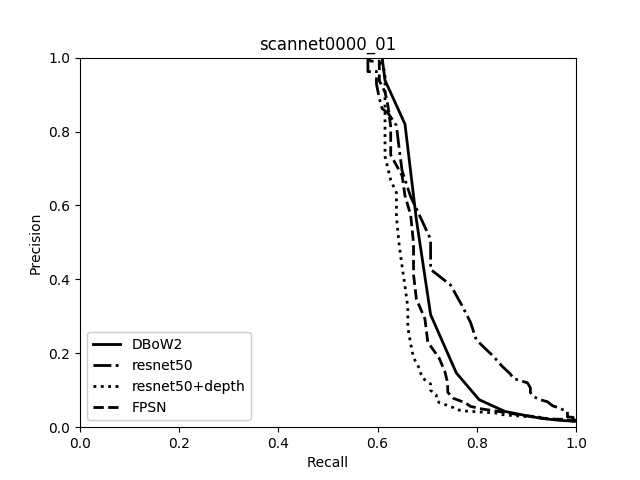} \\
\includegraphics[width=0.4\linewidth]{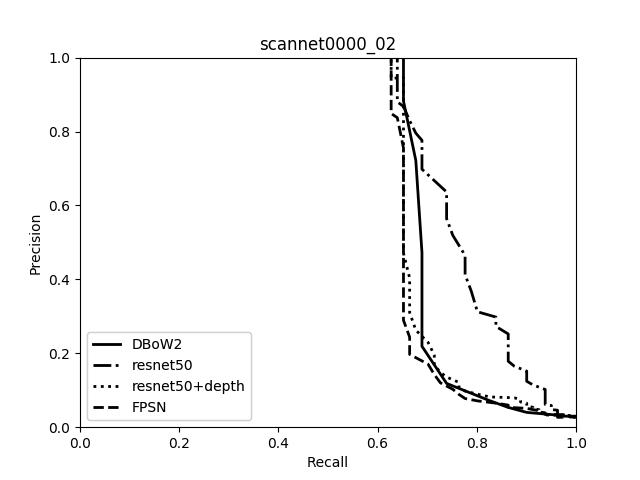} &
\includegraphics[width=0.4\linewidth]{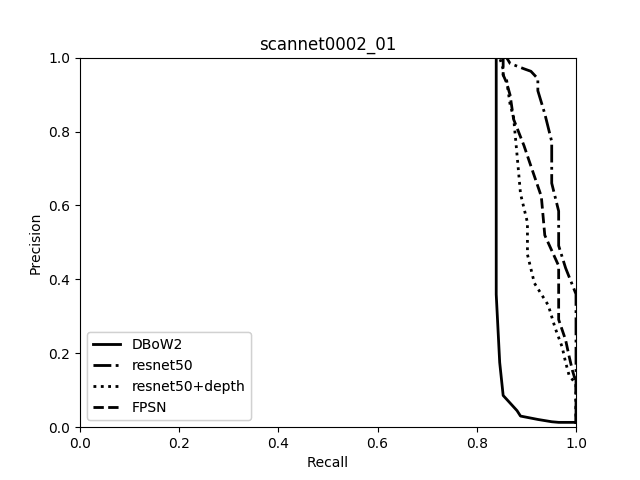}
\end{tabular}
\end{center}
   \caption{Testing results from one Stanford and three ScanNet test areas. Three different networks based on the proposed loop-closure framework are compared against the open-source bag-of-words implementation of DBoW2.}
\label{fig:graph_result}
\end{figure*}

We test three networks on the above datasets: a ResNet50 network trained using only RGB images, a ResNet50 network trained using both RGB and depth images, and the Feature Pyramid Siamese Network (FPSN) proposed in Figure \ref{fig:network} using both RGB and depth images. We compare these networks against the popular open-source implementation for bag-of-words image comparisons DBoW2 \cite{dbow2}. The vocabulary file for DBoW2 is selected as the vocabulary file used in another state-of-the-art solution ORB-SLAM2 \cite{mur2017orb}. We attempted to train the vocabulary file using the Stanford dataset, but it achieved sub par results. Each of the precision-recall curves are shown in Figure \ref{fig:graph_result}. Our network achieves state-of-the-art performance on all test sets. 

On the Stanford 5a dataset, a ResNet50 network trained only with RGB images achieves similar performance as DBoW2, with lower precision than DBoW2 at low recall (below $40\%$ recall), but higher precision at higher recall. We then see further improvement with the addition of depth information, as well as with the use of FPSN.

On the different areas of the ScanNet dataset, while all of our networks still achieve state-of-the-art performance, we see that the ResNet50 with depth information no longer outperforms ResNet50 without depth information. We believe that this is due to the significant difference in depth camera characteristics between our training dataset (collected using high-quality Matterport sensors) and the ScanNet test datasets (collected using portable Occipital Structure sensors). More specifically, the network trained on the Stanford dataset would not have known the depth map characteristics of the ScanNet dataset. We do, however, see FPSN consistently outperform ResNet50, both with depth information, indicating the importance of multi-scale feature detection for loop closure. 

\section{Conclusion and Future Work}
In this paper we have successfully demonstrated the applicability of deep neural networks to the task of pairwise loop closure detection. We show that the inclusion of a depth channel provides new and useful information about the structure of the scene, but may be subject to worse results when the noise of the sensor used for evaluation does not match the noise of the sensor in the training set. Finally we show that the use of our Feature Pyramid Siamese Network architecture improves detection results. Our network achieves the state-of-the-art performance, even outperforming bag-of-words in many cases. We further provide an algorithm for generating training data from large RGB-D datasets, opening the door for further improvements on our results via deep neural networks. 

For this paper, we are only able to find quality RGB-D datasets for indoor environments. As such, we were not able to test our detector on outdoor environments. As depth cameras improve, it will be easier to collect data for a wider set of environments. With such data, it will be possible to apply our algorithm to create an all-purpose loop closure dataset similar to ImageNet that provides a thorough training and testing platform for loop closure.

There are a variety of possible extensions previously proposed for the bag-of-words approach that may allow for improved compute efficiency of this approach. We plan to explore how similar approaches can be developed and applied to deep loop closure detectors enabling these detectors to be used in a wider variety of applications. 

\newpage
{\small
\bibliographystyle{ieee}
\bibliography{loopclosure}
}

\end{document}